\documentclass[a4paper,fleqn]{cas-dc}
\usepackage{graphicx}
\usepackage{subcaption}
\DeclareUnicodeCharacter{01B0}{\u{u}}
\DeclareUnicodeCharacter{01A1}{\u{o}}
\usepackage[authoryear,longnamesfirst]{natbib}
\usepackage{booktabs} 
\usepackage{caption} 
\usepackage[utf8]{inputenc}
\def\tsc#1{\csdef{#1}{\textsc{\lowercase{#1}}\xspace}}
\tsc{WGM}
\tsc{QE}
\tsc{EP}
\tsc{PMS}
\tsc{BEC}
\tsc{DE}

\begin{document}
\let\WriteBookmarks\relax
\def\floatpagepagefraction{1}
\def\textpagefraction{.001}

 \shorttitle{I3D-Based Anomaly Detection in Videos}

\shortauthors{S. Soltani Nejad}

\title [mode = title]{Weakly-Supervised Anomaly Detection in Surveillance Videos Based on Two-Stream I3D Convolution Network}                      



%
\author[1]{Sareh {Soltani Nejad }}[type=editor,
                        orcid=0000-0002-8115-5535]

\cormark[1]


\ead{ssolta7@uwo.ca}



\author[2]{Anwar Haque }

\affiliation[1]{organization={Department of Computer Science},
    addressline={The University of Western Ontario}, 
    city={London, ON},
    postcode={N6A 3K7}, 
    country={Canada}}





\cortext[cor1]{Corresponding author.}



\begin{abstract}
The widespread implementation of urban surveillance systems has necessitated more sophisticated techniques for anomaly detection to ensure enhanced public safety. This paper presents a significant advancement in the field of anomaly detection through the application of Two-Stream Inflated 3D (I3D) Convolutional Networks. These networks substantially outperform traditional 3D Convolutional Networks (C3D) by more effectively extracting spatial and temporal features from surveillance videos, thus improving the precision of anomaly detection. Our research advances the field by implementing a weakly supervised learning framework based on Multiple Instance Learning (MIL), which uniquely conceptualizes surveillance videos as collections of 'bags' that contain instances (video clips). Each instance is innovatively processed through a ranking mechanism that prioritizes clips based on their potential to display anomalies. This novel strategy not only enhances the accuracy and precision of anomaly detection but also significantly diminishes the dependency on extensive manual annotations. Moreover, through meticulous optimization of model settings, including the choice of optimizer, our approach not only establishes new benchmarks in the performance of anomaly detection systems but also offers a scalable and efficient solution for real-world surveillance applications. This paper contributes significantly to the field of computer vision by delivering a more adaptable, efficient, and context-aware anomaly detection system, which is poised to redefine practices in urban surveillance


\end{abstract}




\maketitle
\section{Introduction}
Anomaly detection stands as one of the most intricate challenges within the realm of computer vision \cite{8939930, real, Basharat2008LearningOM, 10.1007/978-981-15-8297-4_53, LIU2020102767, 6531615, inproceedings, Zhao2011OnlineDO}. Specifically, video anomaly detection constitutes a focal point of research, dedicated to discerning uncommon or abnormal behaviors and incidents depicted in video streams. This encompasses a broad spectrum of events, including the unexpected presence of objects or incidents, instances such as a person falling, a vehicular collision, a medical emergency, and deviations from anticipated motion patterns \cite{6531615}.

Anomaly detection remains one of the most challenging and significant areas within the realm of computer vision, particularly in the context of video anomaly detection. This field is crucial for identifying uncommon or abnormal behaviors and incidents depicted in video streams, which include a broad spectrum of events such as unexpected object appearances, medical emergencies, vehicular collisions, and deviations from normal motion patterns [\cite{8939930, real, Basharat2008LearningOM, 10.1007/978-981-15-8297-4_53, LIU2020102767, 6531615, inproceedings, Zhao2011OnlineDO}]. The applications of anomaly detection are diverse and rapidly growing, spanning from medical imaging and traffic monitoring to urban surveillance, where the stakes for timely and accurate detection are particularly high.

Anomaly detection in videos is experiencing rapid growth with diverse applications, including medical imaging, traffic monitoring, and surveillance. Due to the infrequent nature of abnormal occurrences, identifying them manually can be challenging. Therefore, the development of methods to detect patterns deviating from the norm is crucial. Traditional video analysis techniques rely on human monitoring, which is susceptible to errors and time-consuming [\cite{chandola2009anomaly}].

Anomaly detection systems are designed to identify anomalies in video footage and promptly alert users to their presence, proving invaluable in various contexts such as surveillance, industrial monitoring, and other scenarios where detecting abnormalities is critical. The rising demand for urban security has led to an increase in the use of surveillance videos in urban environments to monitor human activity and prevent abnormal events. Detecting anomalies in surveillance videos serves as a crucial tool for security and surveillance, enabling the identification of potential security threats and facilitating prompt action by security personnel. However, continuous monitoring by trained personnel for abnormal events in surveillance videos is both labor-intensive and time-consuming. Hence, research efforts in automatic video anomaly detection are imperative to streamline video monitoring processes, reduce reliance on human resources, and enhance detection accuracy.

To develop a video anomaly detection system, the initial step involves extracting relevant features from the videos. Feature extraction is a critical aspect of this process, requiring the identification and selection of significant patterns and attributes from the video data to differentiate normal from abnormal behavior. This entails analyzing the video data and identifying specific characteristics, such as appearance-based and motion-based features. Appearance-based features, such as color, texture, and shape, are derived from an object's visual appearance and can detect unusual behavior based on changes in object appearance over time. Conversely, motion-based features, derived from the speed, direction, and acceleration of an object's motion, are utilized to detect abnormal movements or sudden changes in motion that may indicate an anomaly.

In recent years, researchers have employed various techniques, including CNNs, VGG architectures, and C3D models, to extract features from visual data. However, a common challenge faced by these methods is their limited capacity to effectively capture temporal features within videos, which are crucial for tasks like anomaly detection. In our study, our primary focus was on addressing this challenge through innovative strategies. Specifically, we investigated the utilization of a two-stream Inflated 3D CNN as our chosen feature extractor. This novel approach facilitates the extraction of both appearance-based (RGB) and motion-based (flow) features from video data. We hypothesized that this integrated approach could yield improved outcomes, as the combination of appearance-based and motion-based features provides a more comprehensive representation of video content. Consequently, we anticipated enhancements in the accuracy and reliability of video anomaly detection systems.

Following feature extraction from video data, these features are utilized to train machine learning models for the accurate detection of anomalous behavior within the video data. Various techniques have been developed for anomaly detection, with recent advancements in deep learning leading to the emergence of new methods tailored for surveillance video analysis. Deep learning methods, particularly, exhibit promising capabilities in outperforming conventional machine learning systems, especially when more data is utilized \cite{phd}. 

Many existing approaches to anomaly detection operate under the assumption that anomalies manifest as deviations from a learned normal pattern. However, this assumption may not always hold true in the context of surveillance videos, which frequently capture complex real-world anomalies that cannot be constructed from normal activities \cite{real}. Moreover, it is impractical to enumerate all conceivable normal activities, as their normality can vary depending on contextual factors \cite{chandola2009anomaly}.

To address the limitations of previous works mentioned earlier, the principal accomplishment of this paper is the implementation of an anomaly detection system capable of detecting anomalies with minimal supervision and less reliance on prior information. The main contributions of this paper can be summed up as follows.

\begin{itemize}
  \item In this paper, we propose a novel approach for video anomaly detection based on a two-stream architecture. Our method employs a two-stream Inflated 3D (I3D) Convolutional Neural Network to extract both RGB and optical flow features from the input video. The RGB stream focuses on capturing information related to object appearance and scene context, while the optical flow stream delineates object motion and dynamics between frames. By integrating the insights from both streams through concatenation of the learned features, our approach achieves a holistic understanding of video content, resulting in enhanced accuracy in anomaly detection.
  \item We extended and refined the anomaly detection model introduced in \cite{real} by leveraging the PyTorch framework \cite{i1}. Our enhanced detector is trained on videos that are weakly labeled as normal or abnormal. To tackle this task, we adopted a weakly-supervised approach based on video-level annotations, employing the Multiple Instance Learning (MIL) framework. Anomaly detection is treated as a regression problem, with the MIL framework utilized to assign elevated anomaly scores to videos anticipated to contain anomalies.
  
 \item We conducted an extensive evaluation of our proposed method on the UCF-Crime dataset \cite{real}, performing experiments to assess its effectiveness in anomaly detection. Our results demonstrate the robust performance of our model, showcasing its capability to accurately detect anomalies in surveillance videos.
\end{itemize}

This paper is organized as follows: In Section 1, we review the existing literature on feature extraction techniques and video anomaly detection methods, establishing the foundation for our research. In Section 2, we present our proposed anomaly detection model, outlining the methodology, including the model architecture and the feature extraction techniques employed. In Section 3, we explore the evaluation metrics and provide experimental results, critically comparing our findings with prior studies. Finally, in Section 4, we conclude with a summary of our contributions and suggest potential avenues for future research to build upon this work.


\section{Related work}
The detection of anomalies in videos poses a significant challenge due to the rarity of abnormal occurrences, making it difficult for human observers to detect such events \cite{phd}. Therefore, the development of methods capable of detecting video patterns deviating from established norms is imperative, termed 'anomaly detection' \cite{chandola2009anomaly}. In recent years, research efforts have intensified to address this challenge within the realm of computer vision \cite{8939930, real, Basharat2008LearningOM, 10.1007/978-981-15-8297-4_53, LIU2020102767, 6531615, inproceedings, Zhao2011OnlineDO, r32, r33}.

Traditional Anomaly Detection Methods:
The foundational approaches in anomaly detection within video surveillance are predicated on the assumption that anomalies are abrupt and rare deviations from normal patterns. Traditional models label these deviations as abnormal events. A variety of statistical models have been employed to capture and encode these normal patterns. Gaussian Process-Based Models utilize Gaussian processes to model and identify deviations in data. Social Force Models, inspired by social dynamics, interpret interactions and movements within crowds. Hidden Markov-based models employ state changes to detect anomalies, adept at handling temporal variations. Histogram-based methods analyze distributions of intensity or color to spot unusual patterns. Motion Patterns focus on analyzing the flow and movement within the video to detect unusual activities. Mixtures of Dynamic Textures Model capture both appearance and dynamics in videos for anomaly detection. Spatial-temporal Markov Random Field Based Models consider both spatial and temporal aspects to identify irregularities. Lastly, Context-Driven Methods incorporate contextual information from the scene to improve detection accuracy.\\
While these methods are effective in controlled settings, they often struggle with scalability and robustness when applied to large-scale video data. The vast volumes and high variability in such datasets pose significant challenges, often leading to decreased performance and reliability in practical surveillance applications.

\textbf{Deep Learning-based Anomaly Detection Methods:} The limitations of traditional anomaly detection techniques have led to the adoption of deep learning algorithms, which are adept at handling complex datasets. These algorithms have significantly enhanced the accuracy and reliability of anomaly detection systems. The transformative impact of deep learning on computer vision has revolutionized the field \cite{r11, r12}. In recent years, deep learning has become the primary method for detecting anomalies in video data. These methods are favored because they train models to recognize abnormalities through detailed learned features, moving beyond the simple deviation metrics used by traditional approaches.
Deep learning approaches to anomaly detection are generally categorized into three main types: reconstruction-based, prediction-based, and hybrid methods. Each category represents a different strategy for modeling and detecting unusual patterns in video data.

\textit{Reconstruction-based methods} are among the most prevalent deep learning techniques utilized in video anomaly detection \cite{r18, r19, r20, r21}. These methods focus on training models to reconstruct video frames. The core principle involves comparing the original frames with their reconstructed counterparts; discrepancies between the two are used to distinguish between normal and abnormal events. Typically, abnormal events generate higher reconstruction errors due to their significant deviation from the learned patterns of normal behavior.

The architecture of models employed for video anomaly detection often mirrors those used in traditional image processing, such as Convolutional Neural Networks (CNNs). However, to effectively capture temporal dynamics critical for video data, these models are enhanced with Long Short-Term Memory Networks (LSTMs) and 3D Convolutional Neural Networks \cite{r7}. Such adaptations allow the models to extend their analytical capabilities beyond static images to dynamic video sequences. Studies utilizing convolutional auto-encoders and deep adversarial training methods have demonstrated the effectiveness of reconstruction-based approaches \cite{r22, r23, r24, r25}.

In the context of video surveillance, significant efforts have been directed towards identifying specific types of anomalies such as instances of violence or aggression. Kooij et al. \cite{n1} leveraged both video and audio data to detect aggressive behaviors. Similarly, Mohammadi et al. \cite{n2} introduced a behavior heuristic-based approach for classifying videos into violent and non-violent categories. Expanding beyond simple classification, some researchers have adopted methodologies that model typical motion patterns of individuals, detecting anomalies by identifying deviations from these patterns \cite{n3}. Techniques such as topic modeling, histogram-based methods, and Hidden Markov Models have been employed to learn these motion patterns without relying on explicit tracking mechanisms, which can often be unreliable \cite{r40, r43, r44, inproceedings}.

While effective in many scenarios, reconstruction-based methods have their limitations. They primarily focus on reconstructing individual frames or patches and do not inherently consider the temporal relationships between successive frames. This limitation can reduce their effectiveness in detecting anomalies that manifest over multiple frames or exhibit unique temporal characteristics that are not apparent in single-frame reconstructions.

\textit{Prediction-based methods} have emerged as a powerful approach for video anomaly detection by utilizing historical data to predict future frames. These methods rely on the premise that deviations from predicted frames signal anomalous events. Auto-encoders are commonly employed to generate these predictions, maintaining consistency with features observed during training \cite{r26, r27}. Notable studies \cite{r28, r29, fewshot} have successfully applied sequence models like Convolutional LSTM (ConvLSTM) to predict future frames and identify discrepancies as potential anomalies.

Additionally, Generative Adversarial Networks (GANs) have been utilized to enhance prediction accuracy and robustness in anomaly detection \cite{r30, gan}. The advantage of prediction-based methods lies in their ability to consider the temporal dynamics of video data, enabling them to detect subtle anomalies that manifest over time—changes that might be overlooked by methods focusing solely on individual frames or short sequences.

Despite their effectiveness, prediction-based approaches, like their reconstruction-based counterparts, often require extensive labeled training data. This need for large volumes of annotated data can be labor-intensive and impractical, particularly for video content where normal behavior varies widely across different contexts. This challenge is compounded by the inherent complexity in defining what constitutes normality, as the same behavior may be deemed normal or abnormal depending on situational context \cite{real}. Furthermore, some complex anomalies may not be effectively captured through these methods because they do not conform to predictable patterns \cite{real}.

To overcome these limitations, there is a growing emphasis on developing anomaly detection techniques that require minimal supervision and are less dependent on prior detailed knowledge of the data. One innovative strategy involves treating anomaly detection as either a classification or a regression problem, utilizing weakly labeled videos. This approach reduces the need for extensive annotations by allowing the model to learn from broadly categorized video data, using anomaly scores to quantify the likelihood of anomalous events. Such methods not only streamline the training process but also enhance the adaptability of the models to diverse and dynamically changing environments.

\textit{Hybrid methods} for detecting anomalies in video content integrate both normal and abnormal videos during the training phase of the anomaly detection model. These approaches leverage a technique known as Multiple Instance Learning (MIL) to facilitate the modeling of motion patterns in a weakly supervised learning environment. In such a setting, the model effectively differentiates between videos containing normal and abnormal events by relying solely on video-level labels, thus reducing the dependency on finely annotated data \cite{r32, r34, r35, real}.

A notable application of this method is the binary classifier developed by Sultani et al., which uses MIL to detect anomalies within video streams \cite{real}. This method benefits from the ability to learn from instances where only the bag (video) labels are available, making it particularly suitable for large-scale video data. Additionally, the incorporation of a graph convolutional neural network, as demonstrated in \cite{r32}, helps to enhance the model's accuracy by cleaning up label noise that typically complicates the training process.

These hybrid methods also address anomaly detection as a regression problem, employing deep-ranking models to calculate anomaly scores. This strategy allows for the quantification of anomalousness in a video sequence, providing a scalable solution for monitoring large volumes of video data in a minimally supervised manner. The utilization of deep-ranking models aids in refining the prediction accuracy by focusing on the relative ranking of anomaly scores rather than absolute binary classifications, thus offering a more nuanced approach to anomaly detection.

Concluding our review of existing anomaly detection methods, we highlight a significant gap in the integration of temporal dynamics with spatial features in video surveillance analysis. Traditional methods, like Convolutional 3D (C3D) models, often focus predominantly on spatial data, neglecting the crucial temporal components necessary for a comprehensive understanding. To bridge this gap, we use the Two-Stream Inflated 3D Convolutional Neural Network (I3D), enhancing detection accuracy by simultaneously analyzing spatial and temporal data. Furthermore, our innovative approach redefines anomaly detection as a regression problem, utilizing a ranking model to precisely score and identify anomalies based on their deviation patterns. 

\section{Methodology}

In this section, we describe the architecture of our two-stream-based video anomaly detection system. We detail the methods utilized to extract both spatial and temporal features from videos, using our advanced two-stream I3D network. Furthermore, we explore our approach to weakly supervised anomaly detection, emphasizing its significant role in enhancing the system’s efficiency and effectiveness in detecting anomalies with minimal manual input.

\subsection{High-Level Architecture}
\begin{figure}
	\centering
		\includegraphics[scale=.20]{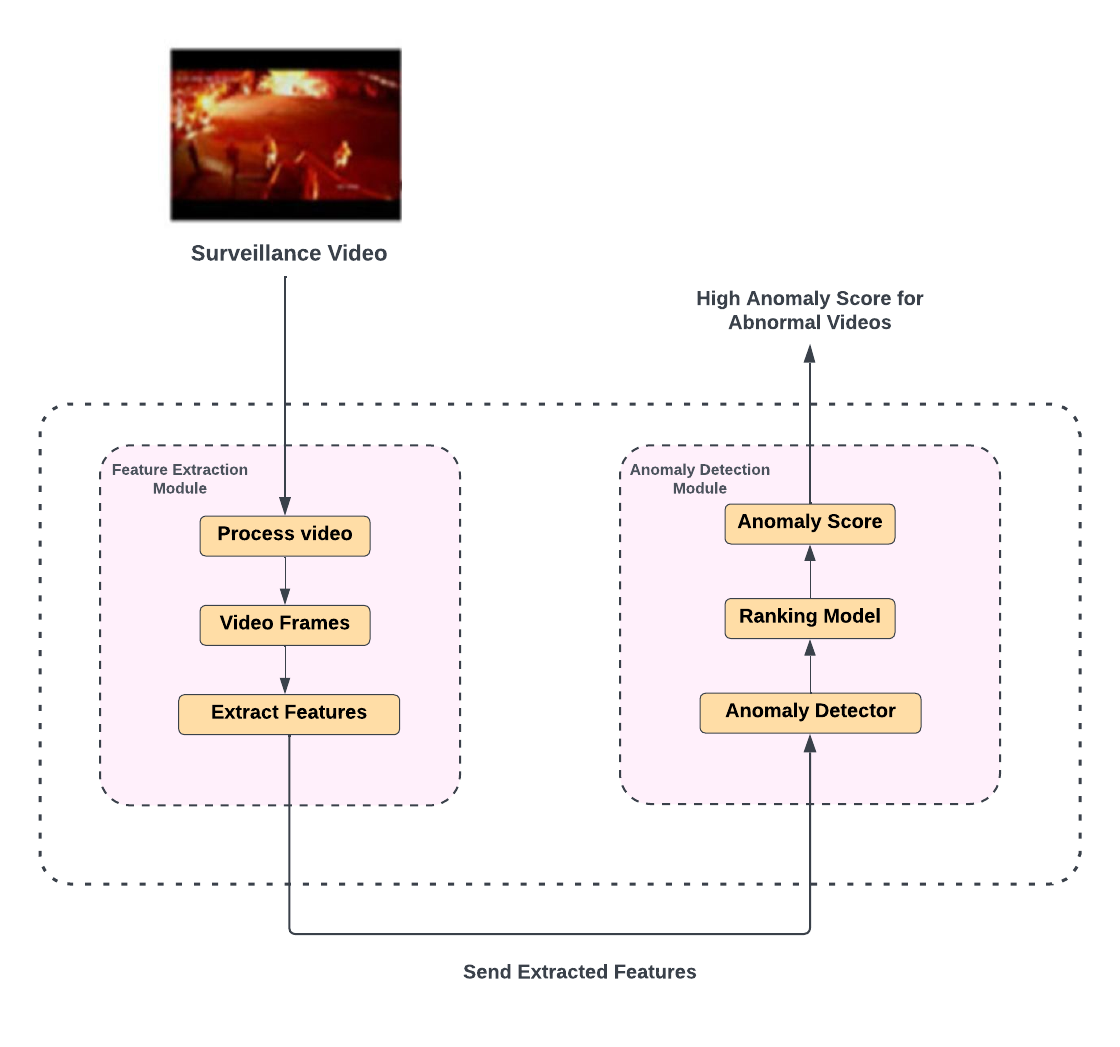}
	\caption{High-Level Proposed System Architecture}
	\label{FIG:1}
\end{figure}

The process of detecting anomalies in surveillance videos within our framework is structured into two distinct stages. Initially, RGB and Flow features are extracted from the videos using a feature extractor module. These features are then utilized in the second stage to detect anomalies, employing a two-stream I3D network for feature extraction and a weakly-supervised anomaly detector based on Multiple Instance Learning (MIL) to identify abnormal behaviors. The overall structure of our framework is depicted in Figure \ref{FIG:1}.

\subsection{Feature Extractor}
The feature extraction pipeline begins by dividing each video into N segments, denoted by \(v_i\) (where i = 1,2,...,N). These segments undergo processing through a two-stream Inflated 3D (I3D) Convolutional Neural Network, trained on the extensive Kinetics 400 dataset (\cite{b6}). This dataset, comprising over 300,000 videos across diverse action categories, offers a robust foundation for model training and feature learning (\cite{b6}).

\begin{figure*}
\centering
\includegraphics[scale=.36]{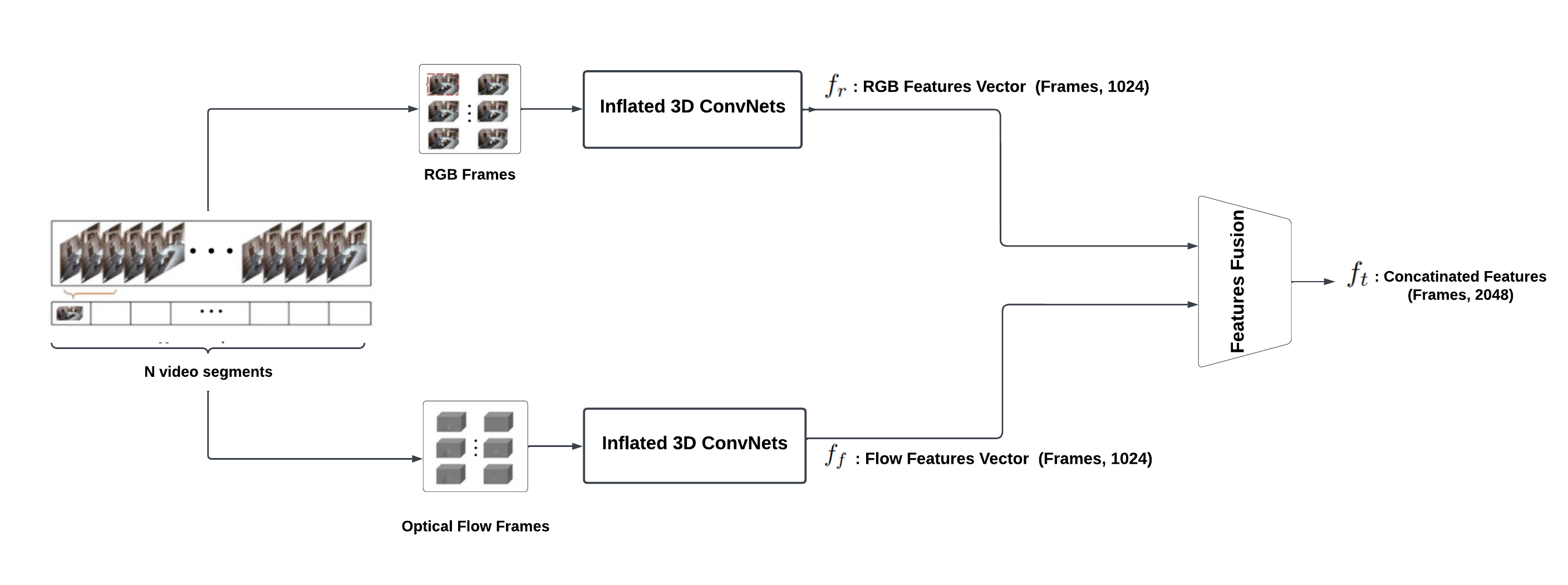}
\caption{Architecture of the two-stream Inflated 3D Convolutional Neural Networks}
\label{FIG:2}
\end{figure*}

The I3D architecture marks a significant advancement from conventional 2D Convolutional Neural Networks, as it adeptly captures both spatial and temporal dimensions, essential for robust feature extraction from intricate video data (\cite{r10}). This holistic understanding allows the model to discern not only the appearance of objects but also their dynamic movements within the frames.

In our approach, we employ two parallel Inflated 3D ConvNet (I3D) networks, as illustrated in \ref{FIG:2}, with one network trained exclusively on RGB frames and the other on optical flow data. This dual-network configuration significantly enhances the model’s ability to interpret complex video content \cite{b8}. 

The RGB network processes video frames through a series of convolutional and pooling layers, effectively capturing both spatial and temporal dimensions. The resultant output is a 1024-dimensional tensor, designated as \(f_r\), which contains detailed visual information including object positioning, shape, lighting variations, and color attributes.
 
In parallel, the optical flow network employs the TVL1 algorithm \cite{b10} to calculate the motion between consecutive frames, achieving exceptional smoothness and accuracy. This motion data is then processed by additional convolutional layers, culminating in the generation of a 1024-dimensional tensor, designated as \(f_f\), which provides profound insights into object movements and activities within the video scene

RGB and Flow features provide complementary information that can be synergistically merged to enhance video content representation. In this integration process, the outputs from the two distinct streams are combined during a late fusion phase, which involves concatenating the learned features to generate a unified prediction. Specifically, the two-stream Inflated 3D ConvNet (I3D) generates two separate 1024-dimensional tensors—one representing RGB features and the other representing Flow features. These tensors are subsequently concatenated to form a comprehensive 2048-dimensional tensor that encapsulates both feature sets. This concatenated tensor, denoted as \(f_t\), serves as the input to the anomaly detection model, which is further elaborated in the subsequent section.

\subsection{Weakly-Supervised Anomaly Detection Model}
\begin{figure*}
\centering
\includegraphics[scale=.28]{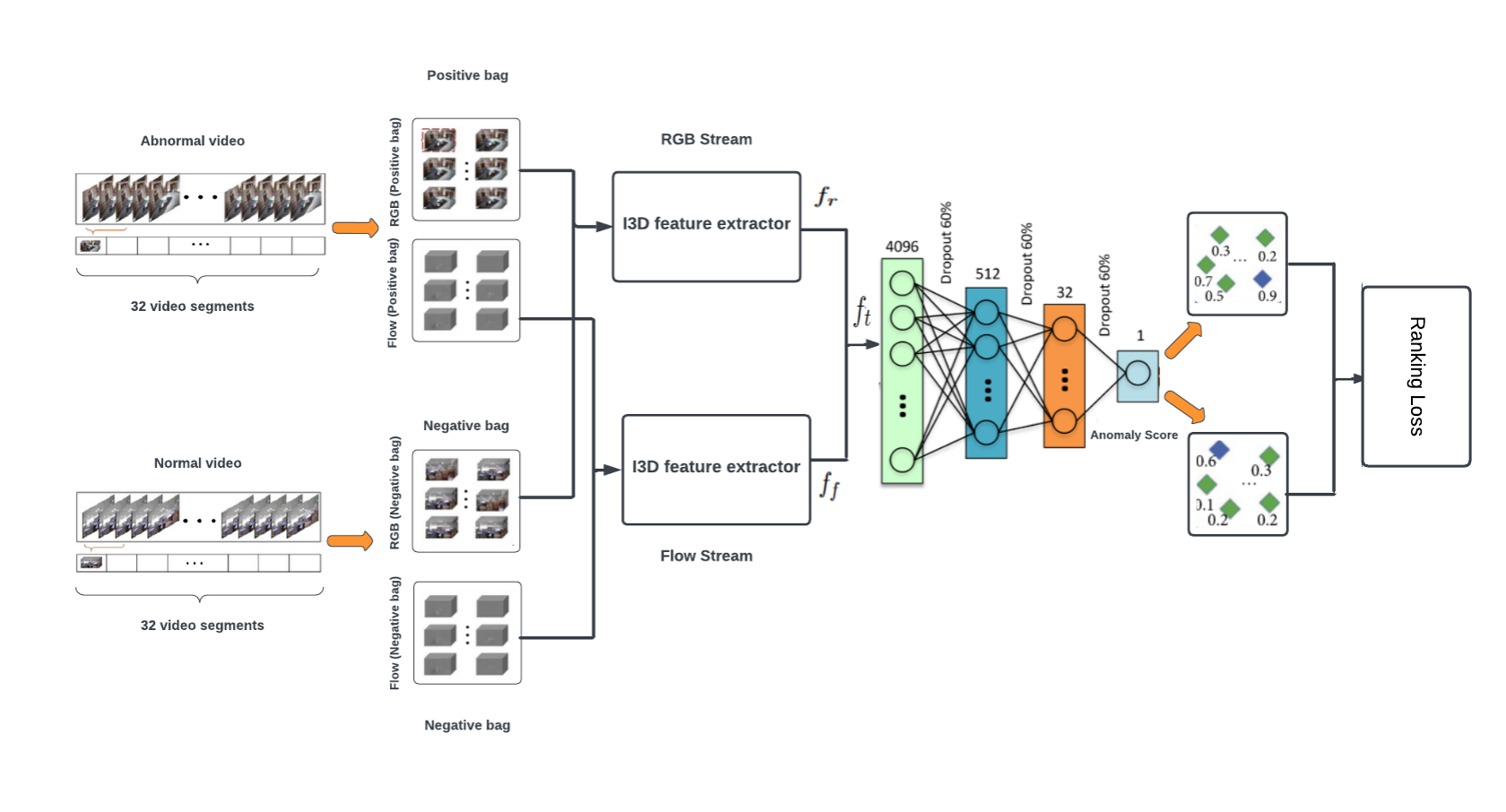}
\caption{The architecture of the proposed anomaly detection model}
\label{FIG:3}
\end{figure*}

 Our paper introduces a Weakly-Supervised Anomaly Detection Model that utilizes concatenated feature vectors, denoted as \(f_t\). These vectors, which integrate both RGB and Optical Flow features from video segments, serve as the primary inputs for our anomaly detection framework, as depicted in Figure~\ref{FIG:3}. Our model builds on the anomaly detection framework originally proposed in \cite{real} and utilizes PyTorch \cite{i1} for enhancement, specifically tailored to manage the ambiguity inherent in weakly labeled surveillance video data. 

Traditional anomaly detection methods typically require detailed instance-level annotations. In contrast, our approach leverages video-level labels, classifying videos as normal or abnormal without specifying the exact locations of anomalies. This classification significantly reduces the need for manual labeling, aligning with the principles of weakly-supervised learning.

To effectively interpret these broad categorizations, we integrate the Multiple Instance Learning (MIL) framework \cite{b2}. MIL is an approach where training data are grouped into "bags," each assigned a single class label. The individual instances within these bags, which may not be labeled, determine the bag's classification: a bag is positive (\(B_a\) if it contains at least one positive instance, and negative (\(B_n\) if all instances are negative.

To apply the MIL framework to our anomaly detection challenge, we segmented each video into 
N parts, each defined as \(v_i\) (where i = 1,2,...,N), and each video was represented by \(V = \{v\}_i^N\). Therefore, we considered each video as a bag and each video segment as an instance within the bag. Videos with at least one anomalous segment are labeled as positive bags (\(B_a\), while those without anomalies are considered negative bags (\(B_n\). We then extracted I3D features from each segment using a two-stream I3D network, shown in Figure~\ref{FIG:2}. These features were input into a three-layer fully connected neural network to assign an anomaly score to each segment, enhancing the model's ability to detect anomalies efficiently and accurately in real-world scenarios.

\textbf{Ranking Model:}  In addressing the challenges posed by the limited availability of abnormal videos for training and the time-consuming task of annotating segment-level labels, we adopted a regression-based approach to anomaly detection, rather than a traditional classification framework.

We implemented a deep ranking loss mechanism, as proposed in \cite{real}, to prioritize higher anomaly scores for abnormal segments over normal segments. This prioritization is formally expressed as:
\begin{equation} \label{e1}
\Large f(v_a) > f(v_n)
\end{equation} 
Here, \(f(v_a)\) represents the predicted anomaly score for an abnormal video segment, while 
\(f(v_n)\) denotes the score for a normal segment. Given the absence of explicit segment-level annotations for videos, we calculate the ranking loss by comparing the highest scores within the positive bag (\(B_a\)), which contains segments with anomalies, to those in the negative bag (\(B_n\)), which comprises segments without anomalies. This modified ranking function is given by:

\begin{equation} \label{e2}
\Large \max_{i\epsilon B_a}f(v_a^i) > \max_{i\epsilon B_n} f(v_n^i)
\end{equation}  
where \(f(v)\) denotes the predicted score for given video segment.   
The purpose of this approach is to train the anomaly detector to distinguish between abnormal and normal segments even in the absence of explicit annotations.\\
The ranking loss between the top-scoring instances in the positive bag and the negative bag is computed using the hinge-loss function, as stated in \cite{real}. The formula for the hinge-loss function is derived from equation \ref{e2} and is as follows: 
\begin{equation} \label{e3}
\Large l(B_a,B_n) = \max (0, 1 - \max_{i\epsilon B_a}f(v_a^i) + \max_{i\epsilon B_n} f(v_n^i)
\end{equation} 
To account for the typically brief duration of anomalies, we incorporate a sparsity constraint into our loss function, ensuring that high anomaly scores remain sparse. Additionally, a smoothness constraint is applied to ensure gradual changes in anomaly scores across adjacent segments, reflecting the sequential nature of video data. The comprehensive loss function, integrating both constraints, is detailed as follows:
\begin{equation}
\Large l(B_a,B_n) = \max (0, 1 - \max_{i\epsilon B_a}f(v_a^i) + \max_{i\epsilon B_n} f(v_n^i)  
\end{equation} 
\begin{equation} \label{e4}
+ \Large \lambda_1\sum_{i}^{n} f(v_a^i) + \lambda_2\sum_{i}^{n-1} (f(v_a^i) - f(v_a ^ (i+1)))^2 
\end{equation} 
Where $\lambda_1 $ and  $\lambda_2 $ are the sparsity and smoothness parameters, respectively, set to 0.00008 as recommended in \cite{real}.

\section{Experiments}
\subsection{Dataset}
Our model is evaluated on the UCF-Crime dataset, created specifically for the weakly supervised video anomaly detection task \cite{real}. This dataset comprises an extensive collection of 1900 lengthy, Untrimmed, real-world surveillance videos, totaling 128 hours. The UCF-Crime dataset is particularly notable for its balanced composition, including 950 videos with real-world anomalies and 950 normal videos, which facilitates a comprehensive evaluation of anomaly detection models.

The dataset is structured into 13 realistic anomaly types designed to closely mimic real-life situations, including Abuse, Arrest, Arson, Assault, Accident, Burglary, Explosion, Fighting, Robbery, Shooting, Stealing, Shoplifting, and Vandalism. This diversity makes the dataset especially suitable for training and evaluating models intended to detect sophisticated anomalies in surveillance settings. The training set consists of 800 normal videos and 810 abnormal videos, while the testing set comprises 150 normal and 140 abnormal videos, ensuring coverage of all anomaly categories.

In our experiments, we utilized the default class mapping provided by the UCF-Crime dataset to facilitate comparison with baseline results. This extensive dataset provides a robust foundation for testing our model's ability to recognize diverse and intricate anomalies within complex environments.

\subsection{Implementation Details}
We developed our anomaly detection model using PyTorch \cite{i1}. The model leverages a two-stream Inflated 3D Convolutional Neural Network (I3D) \cite{r10}, originally trained on the Kinetics dataset, to extract visual features from surveillance videos. Each video was divided into 32 segments, with each segment consisting of 16 consecutive frames. The I3D network processed each segment to extract RGB and Flow features, culminating in two 1024-dimensional tensors for each frame. These tensors represent the RGB features, detailing visual appearance aspects such as color, texture, and shape, and Flow features, which provide insights into object motion dynamics including direction, speed, and acceleration.

To synthesize a comprehensive feature set for each segment, we computed the average of the extracted features across all 16 frames. These averaged features from both RGB and Flow streams were then concatenated, forming a 2048-dimensional tensor that served as input to our neural network model.

The neural network architecture includes three fully connected (FC) layers. The first layer contains 512 units, designed to condense the high-dimensional input into more manageable representations. It is followed by a second layer with 32 units, further refining the features, and a final output layer consisting of a single unit. This architecture was adapted from the baseline anomaly detector proposed in \cite{real}.

In our experimental setup, the model was trained using batches of 30 videos, each randomly selected from both the abnormal and normal video subsets of the UCF-Crime dataset.  We utilized the Adagrad optimizer \cite{i2} for optimization, with a learning rate set at 0.001. This configuration was chosen to ensure efficient convergence while maintaining a balance between the speed and accuracy of the learning process.

\subsection{Evaluation metric}
To assess the performance of our anomaly detection model, we use the frame-based Receiver Operating Characteristic (ROC) curve and its corresponding Area Under the Curve (AUC) as the primary evaluation metric. This metric is not only widely recognized but also the most commonly used in previous research on video anomaly detection \cite{real, c1, c2, r26}. This metric allows us to evaluate the effectiveness of our model and make comparisons with other state-of-the-art methods in the anomaly detection domain. 

\subsection{Experimental Results}
\begin{figure*}[t]
    \centering
    \begin{subfigure}[b]{0.49\textwidth}
        \includegraphics[width=\textwidth]{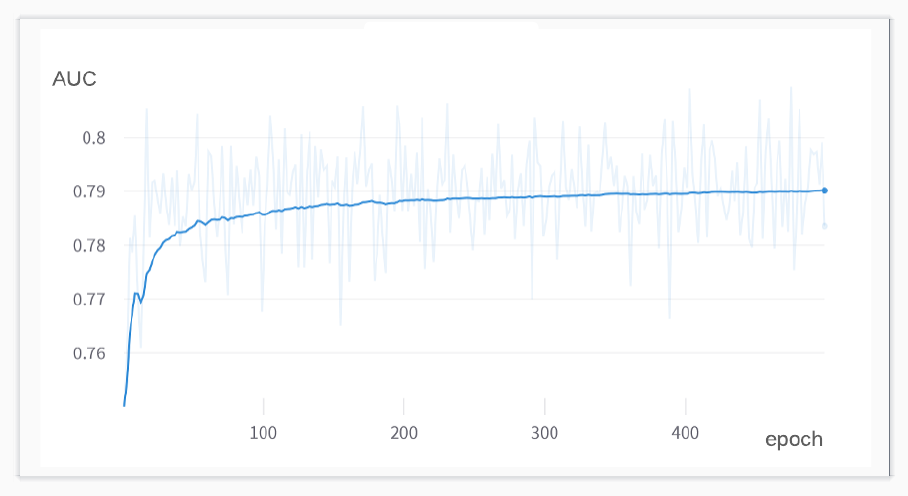}
        \caption{The AUC results based on I3D RGB stream network}
        \label{FIG:4}
    \end{subfigure}
    \hfill 
    \begin{subfigure}[b]{0.49\textwidth}
        \includegraphics[width=\textwidth]{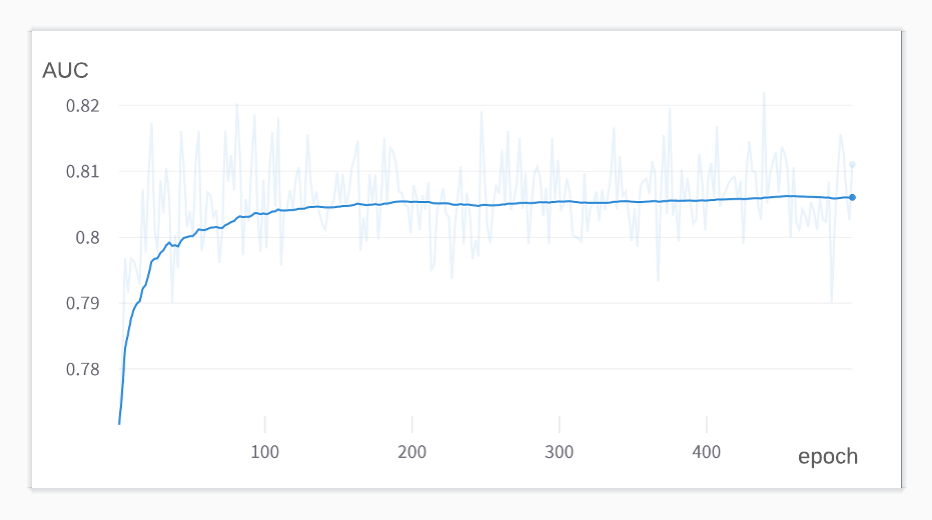}
        \caption{The AUC results based on I3D Flow stream network}
        \label{FIG:5}
    \end{subfigure}
    \vspace{1cm} 
    \begin{subfigure}[b]{0.49\textwidth}
        \includegraphics[width=\textwidth]{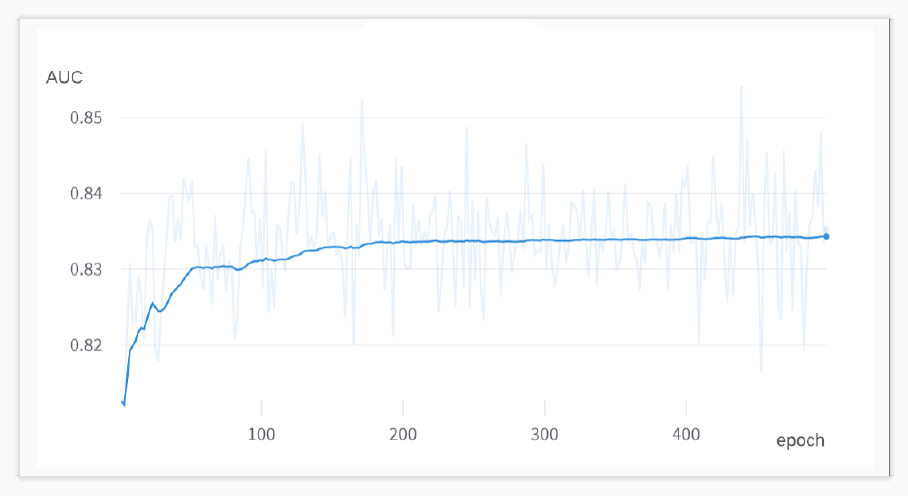}
        \caption{The AUC results based on two-stream I3D network}
        \label{FIG:6}
    \end{subfigure}
    \hfill 
    \begin{subfigure}[b]{0.49\textwidth}
        \includegraphics[width=\textwidth]{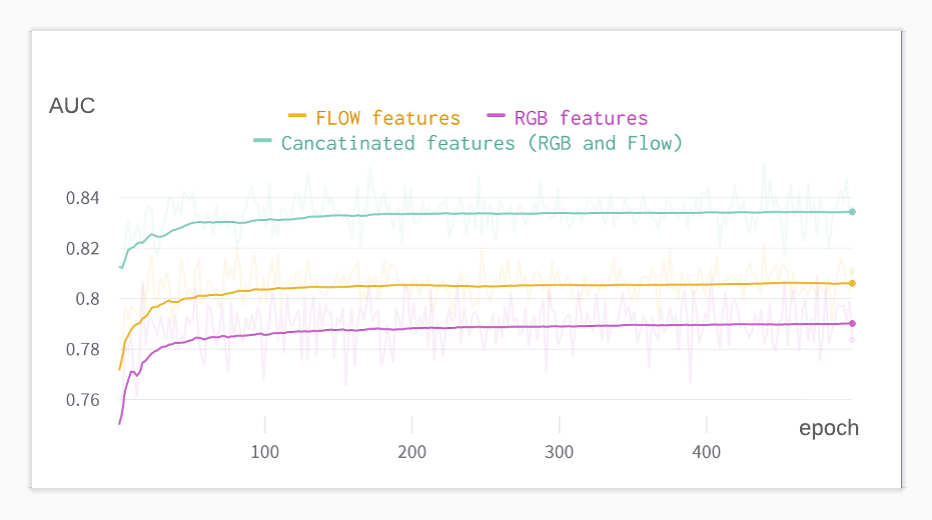}
        \caption{Comparison of the AUC results for the three scenarios}
        \label{FIG:7}
    \end{subfigure}
    \vspace{-1cm}
    \caption{Comparative AUC results across different feature extraction models on the UCF-Crime dataset, illustrating the performance enhancements achieved through I3D RGB, I3D Flow, and two-stream network configurations.}
\end{figure*}

In this section, we evaluate the performance of our anomaly detection model across different types of video feature extraction networks by measuring the Area Under the Curve (AUC) in three distinct scenarios. Initially, the model's performance was assessed using the I3D RGB stream network, with the results depicted in Figure~\ref{FIG:4}, focusing on the model’s ability to detect anomalies based solely on RGB data.

Subsequently, the model's effectiveness in detecting anomalies was analyzed using the I3D flow stream network, which concentrates on motion-based features, with outcomes illustrated in Figure~\ref{FIG:5}. The third evaluation involved the two-stream network that integrates both RGB and flow streams, with the corresponding AUC results presented in Figure~\ref{FIG:6}. 

Further comparative analysis of these scenarios is provided in Figure~\ref{FIG:7}, which presents a comparison of the AUC results based on the UCF-Crime dataset. This analysis clearly demonstrates that the anomaly detection model utilizing the two-stream I3D network significantly outperforms those using individual RGB or Flow streams. The superior performance of the two-stream model indicates that combining RGB and Flow features provides a more comprehensive representation of video content, thereby enhancing the model’s capability to effectively detect anomalies in surveillance videos. This combined approach leverages the distinct advantages of each feature type, proving particularly advantageous for detecting anomalies in complex surveillance scenarios.

\begin{figure}[h]
\centering
\includegraphics[scale=0.22]{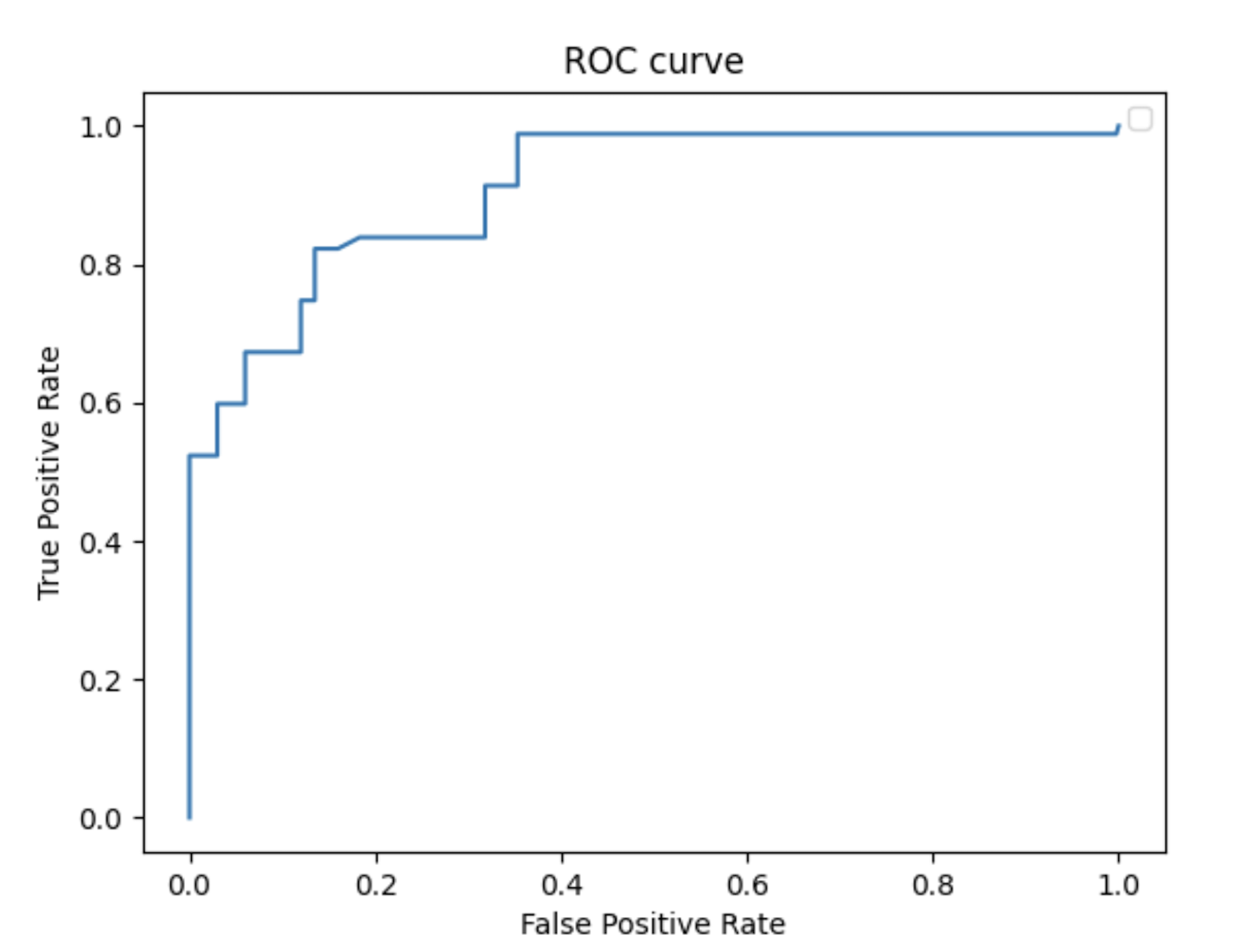}
\caption{ROC Curve of the proposed anomaly detection method, illustrating the trade-off between the True Positive Rate (TPR) and False Positive Rate (FPR) across various threshold settings.}
\label{Fig:8}
\end{figure}

Building on the comparative analysis of the AUC results from the three scenarios, the Receiver Operating Characteristic (ROC) Curve for the proposed system further elucidates its performance. Illustrated in Figure \ref{Fig:8}, the ROC Curve displays the relationship between the True Positive Rate (TPR) and the False Positive Rate (FPR). This visualization is crucial as it effectively demonstrates the system's ability to discriminate true positive detections from false positive instances, providing an additional layer of validation for the model's effectiveness in anomaly detection. The ROC curve complements the findings discussed previously, confirming the superior discriminative power of the two-stream network as evidenced by the AUC results in Figure \ref{FIG:7}.

\begin{figure}[h]
\centering
\includegraphics[scale=1]{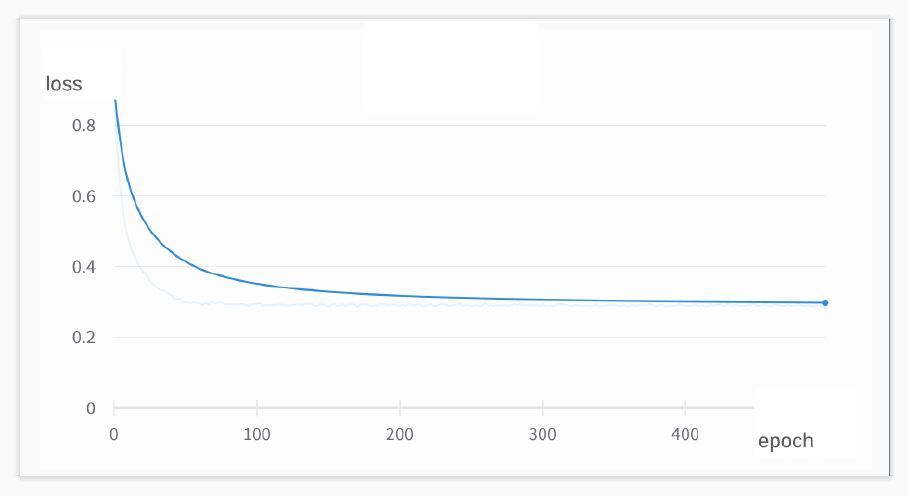}
\caption{Training loss curve of the two-stream network, illustrating a decreasing trend in loss over successive training epochs, which signifies improving model performance.}
\label{Fig:9}
\end{figure}

Figure~\ref{Fig:9} presents the model loss incurred during the training phase of the two-stream network. As depicted, the loss decreases over time, indicating the model’s improving performance as training progresses.

\subsection{Comparison with the State-of-the-art}
We conducted a comparative analysis of our anomaly detection approach against established methods in the field. Hasan et al. \cite{c1} developed a method utilizing a fully convolutional feed-forward deep autoencoder that learns local features to train a classifier. Lu et al. \cite{c2} proposed a dictionary-based approach that learns patterns of normal behavior and employs reconstruction errors to identify anomalies. Sultani et al. introduced a model that leverages C3D features for anomaly detection, which we selected as our baseline for comparison.

Quantitative comparisons were performed using the AUC metric on the UCF-Crime dataset, as detailed in Table \ref{table:1}. The results indicate that our proposed method, which utilizes the two-stream I3D network for feature extraction, surpasses the performance of the baseline model by Sultani et al. \cite{real}. This improvement is clearly reflected in the AUC values presented in the table, confirming the superior efficacy of our approach in detecting anomalies in surveillance video contexts.

\renewcommand{\arraystretch}{1.4} 
\setlength{\tabcolsep}{23pt} 
\begin{table}[ht]
\centering
\caption{Quantitative comparison of anomaly detection methods on the UCF-Crime dataset.}
\label{table:1}
\begin{tabular}{@{}lc@{}}
\toprule
\textbf{Method}                        & \textbf{AUC (\%)} \\
\midrule
Hasan et al.                           & 50.6              \\
Lu et al.                              & 65.51             \\
Sultani et al. (C3D)                   & 75.41             \\
Ours (I3D RGB)                         & 80.93             \\
Ours (I3D Flow)                        & 82.20             \\
Ours (I3D RGB \& Flow)                 & 85.41             \\
\bottomrule
\end{tabular}
\end{table}

\subsection{Hyper-parameter Tuning}
As part of our model optimization process, we engaged in a systematic search for the most effective optimizers and learning rates, focusing on two widely used optimization algorithms in deep learning: Adam \cite{i3} and Adagrad \cite{i2}. We evaluated these optimizers across a set of predefined learning rates, specifically [0.01, 0.001, 0.0001], to determine which combination yields the best performance.

\begin{figure}[hbt!]
\centering
\includegraphics[scale=0.5]{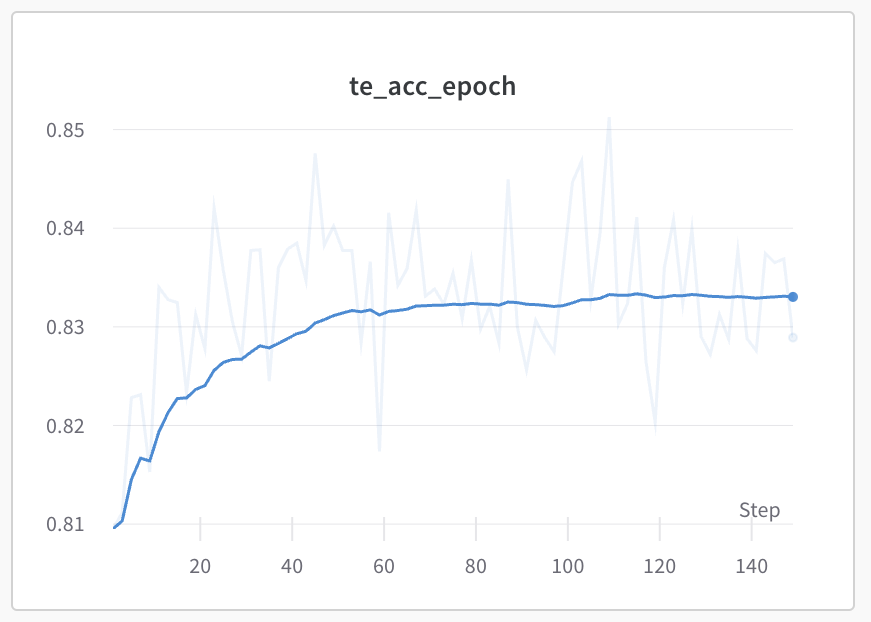}
\caption{\small AUC results demonstrating the performance of the Adagrad optimizer across various learning rate settings}
\label{fig: adagrad}
\end{figure}

\begin{figure}[hbt!]
\centering
\includegraphics[scale=0.5]{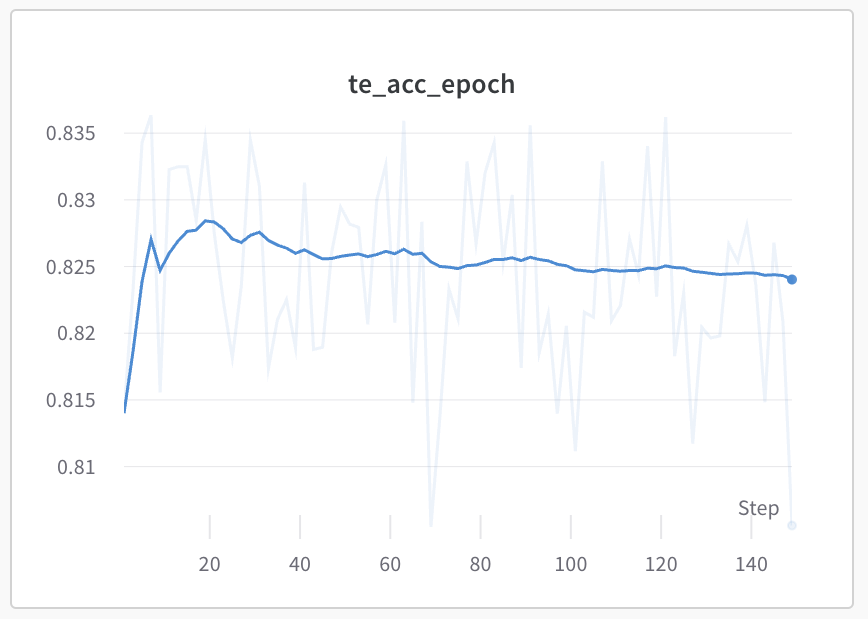}
\caption{\small{AUC results showcasing the efficacy of the Adam optimizer under different learning rate scenarios.}}
\label{fig: adam}
\end{figure}

The comparative results, displayed in Figure \ref{fig: adagrad} for Adagrad and Figure \ref{fig: adam} for Adam, indicate that Adagrad significantly outperforms Adam in the context of anomaly detection within our model. We attribute this enhanced performance to Adagrad's lower operational complexity and its method of adapting the learning rate on a per-parameter basis. Unlike Adam, Adagrad modifies each parameter's learning rate based on the historical sum of its gradients. This feature is particularly advantageous in scenarios involving sparse data or when the model has to navigate high-dimensional feature spaces, as it promotes a more tailored and hence efficient optimization process.

Overall, the findings from our hyperparameter tuning underscore Adagrad's suitability as an optimizer for anomaly detection tasks in complex deep learning models, suggesting its potential superiority over Adam in such applications.

\section{Conclusion}
In this paper, we tackled the complex challenge of detecting anomalies in video surveillance, where traditional methods typically underperform. Traditional anomaly detection frameworks often operate under the assumption that anomalies manifest as deviations from learned normal behaviors. This assumption, however, is inadequate for surveillance videos that capture diverse and intricate real-world anomalies, which do not necessarily follow predictable patterns. To address this gap, we developed a novel methodology that reduces reliance on extensive prior knowledge. Our approach utilizes a Two-Stream Inflated 3D Convolutional Neural Network (I3D) to extract both RGB and Flow features from video data. Integrating these dual streams allows our model to achieve a deeper, more nuanced understanding of video content, significantly enhancing anomaly detection accuracy.
We evaluated our method on the UCF-Crime dataset. The results unequivocally demonstrate that our model outperforms existing approaches in identifying anomalies in surveillance footage. These findings not only confirm the effectiveness of our approach but also open new avenues for future research aimed at advancing anomaly detection technologies.

\section{Limitations and Future Work}
The proposed anomaly detection system, while high-performing, incurs significant computational costs, especially when compared to simpler models like C3D. This is largely due to the Two-Stream Inflated 3D Convolutional Neural Network (I3D) model, which processes each video frame twice—once per stream—to integrate both spatial and temporal information, thus demanding more computation and memory.

Additionally, operating under a weakly supervised framework reduces the need for detailed annotations but increases the likelihood of false positives, particularly in low-light conditions. To address these challenges, enhancing the training dataset with labeled anomalies or integrating auxiliary data sources such as thermal imaging or audio signals could improve the accuracy and robustness of the system.

\bibliographystyle{cas-model2-names}

\bibliography{references}


\end{document}